# Intelligent Three-Level Learning Architecture for Autonomous UAV Swarms in Search-and-Rescue Operations

Oleksii Bychkov
*Department of Program Systems and Technologies*
*Taras Shevchenko National University of Kyiv*
*Kyiv, Ukraine*
oleksiibychkov@knu.ua, ORCID: 0000-0002-9378-9535

***Abstract*—This paper presents a novel three-level hierarchical learning architecture for autonomous UAV swarms performing search-and-rescue (SAR) operations. Unlike conventional approaches that apply a single learning paradigm across all hierarchy levels, the proposed architecture integrates three qualitatively different learning mechanisms corresponding to the biological hierarchy of reflexes, skills, and reasoning: Hebbian neuroplasticity for individual agent adaptation (Level 1), multi-agent reinforcement learning with graph neural networks and behavior trees for tactical coordination (Level 2), and model-agnostic meta-learning with BDI reasoning and a digital twin for strategic decision-making (Level 3). The architecture is formalized through a system of 22 architectural contracts organized across six components (BDI, Behavior Trees, GNN, MARL, Neuroplasticity, Meta-Learning) that collectively provide six classes of formal guarantees: safety, budget correctness, optimality, liveness, starvation-freedom, and inter-level consistency. We introduce the concept of Swarm Meta-Cognition (SMC) as a compositional property arising from the structured interaction of all three levels, enabling the swarm to monitor its own cognitive state and switch between cognitive strategies. Five constructive progress functions for SAR task types (coverage, verification, tracking, delivery, relay) bridge the gap between the abstract optimization theory and concrete operational scenarios. The main integration theorem (Theorem HAF) establishes that when all contracts are satisfied, the hybrid neuro-symbolic system preserves all six guarantee classes. For the dynamic case with active learning at all levels, five new contracts (NP-C1–C3, ML-C1–C2) extend the framework with three additional guarantees: cognitive resilience, graceful degradation, and monotonic meta-improvement. Theoretical analysis demonstrates that the proposed architecture addresses five fundamental limitations of existing hierarchical RL approaches.**

## I. INTRODUCTION

Unmanned aerial vehicles (UAVs) have undergone a fundamental transformation from individually operated reconnaissance platforms to autonomous swarms capable of executing complex missions. This evolution is driven by the convergence of affordable microelectronics, advances in artificial intelligence (particularly deep learning and reinforcement learning), and the operational requirement for rapid decision-making beyond human reaction times. Search-and-rescue (SAR) operations represent a particularly demanding application domain, requiring coverage of large areas, real-time victim detection and verification, dynamic task reallocation under uncertainty, and resilience to communication disruptions and agent failures.

Recent conflicts, including the full-scale war in Ukraine (since 2022), has demonstrated that swarms of small UAVs can achieve strategic effects that were previously the domain of expensive manned platforms. However, current operational swarms rely heavily on human operators in the control loop, creating bottlenecks in decision-making speed and limiting scalability. The transition from Level 4 autonomy (supervised autonomy with human approval for critical decisions) to Level 5 (full autonomy with self-organized decision-making, meta-learning, and swarm meta-cognition) requires fundamentally new types of intelligence that existing approaches do not provide.

Current approaches to UAV swarm intelligence fall into two broad categories. The first relies on centralized control with homogeneous agents, which creates a single point of failure, limits scalability, and provides no graceful degradation under communication loss. The second employs decentralized multi-agent reinforcement learning (MARL), which provides adaptability but typically lacks formal safety guarantees, suffers from non-stationarity when all agents learn simultaneously, and cannot bridge the gap between reactive behavior and strategic reasoning.

A critical analysis of existing hierarchical RL for multi-agent systems reveals five fundamental limitations that motivate this work. First, single-type learning: all hierarchy levels (e.g., manager and worker in Feudal Networks, both levels in h-QMIX) use the same gradient-based mechanism, meaning a distribution shift requires complete retraining at all levels. Second, non-stationarity: simultaneous learning by multiple agents creates non-stationary environments that violate standard convergence assumptions. Third, the neuro-symbolic integration problem: how to connect BDI-style symbolic reasoning with learned neural policies under formal guarantees remains unsolved. Fourth, no existing architecture provides constructive safety guarantees for hybrid systems with neural components. Fifth, the relationship between individual adaptation and collective performance is not formalized, making it impossible to bound how changes in one agent propagate to swarm-level behavior.

This paper proposes a three-level hierarchical learning architecture that addresses all five limitations by assigning qualitatively different learning mechanisms to each level, mirroring the biological hierarchy from reflexes through skills to reasoning. The contributions of this paper are: (1) a biologically inspired architecture with Hebbian neuroplasticity at Level 1 (individual adaptation), MARL with GNN coordination at Level 2 (tactical coordination), and MAML-based meta-learning with BDI reasoning at Level 3 (strategic decision-making); (2) a formal contractual framework of 22 architectural contracts providing six classes of guarantees; (3) five constructive progress functions for SAR task types that bridge abstract theory and concrete operations; (4) the concept of Swarm Meta-Cognition (SMC) as a compositional property; and (5) analysis of graceful degradation under progressive communication loss.

The remainder of this paper is organized as follows. Section II presents the system model and mathematical notation. Section III describes the three-level learning architecture in detail. Section IV develops the mathematical framework including progress functions and optimization. Section V presents the contractual framework with 22 contracts and six guarantee classes. Section VI analyzes inter-level integration and graceful degradation. Section VII defines Swarm Meta-Cognition. Section VIII provides comparison with existing approaches. Section IX discusses theoretical limitations and future directions. Section X concludes the paper.

## II. SYSTEM MODEL AND NOTATION

### A. Swarm Structure

The swarm S is defined as a three-level hierarchical structure: $\mathcal{S} = \langle \mathcal{L}_{\text{cmd}}, \{\mathcal{G}_g\}_{g=1}^{G}, \{\mathcal{A}_i\}_{i=1}^{N} \rangle$, where $\mathcal{L}_{\text{cmd}}$ is the Swarm Executive (decentralized command center), $\mathcal{G}_g$ is the commander of functional group $g$ $(g = 1, \dots, G)$, and $\mathcal{A}_i$ is agent $i$ $(i = 1, \dots, N)$. The Swarm Executive is realized in a distributed manner through three mechanisms:

(a) a replicated TaskBoard with eventual consistency, where each group maintains a local copy of the global task state synchronized via CRDT (Conflict-free Replicated Data Types);

(b) an auction mechanism with local decisions, where task allocation occurs through competitive bidding without a centralized arbiter;

(c) a leader election protocol for network partitions, so that each disconnected subgroup elects its own local Executive.

### B. Task Model

The mission is decomposed into tasks from a finite type set: $\Upsilon = \{\text{coverage, verification, tracking, delivery, relay}\}$ . Each task $j$ has a type $\upsilon_j \in \Upsilon$ that determines the shape of its progress function. Tasks are further classified as ground tasks $\mathcal{T}^{\text{gr}}$ (originating from the mission operator) and self-generated tasks $\mathcal{T}^{\text{sg}}$ (discovered by the swarm during operation), with $\mathcal{T}_k = \mathcal{T}_k^{\text{gr}} \cup \mathcal{T}_k^{\text{sg}}$ at each time step $k$. This distinction is critical for the starvation-freedom guarantee: ground tasks receive prioritized allocation through a separation constant $\lambda$.

### C. Control Variables and Constraints

The system operates with two levels of control variables. At the group level, $y_{gj}(k) \geq 0$ represents the resource allocated by group $g$ to task $j$ at time step $k$. At the agent level, $x_{ij}(k) \in \{0,1\}$ indicates whether agent i participates in task $j$. The two levels are linked by the aggregation relation: $y_{gj}(k) = \sum_{i \in \mathcal{A}_g} c_i \cdot x_{ij}(k)$ , where $c_i$ is the capacity of agent i (a scalar combining flight time, energy, and sensor capability).

Two fundamental constraints govern the system. The group budget constraint ensures that the total resource allocated by group $g$ does not exceed its available budget: $\sum_j y_{gj}\,(k) \leq B_g(k)$. The single-assignment constraint ensures that each agent works on at most one task: $\sum_j x_{ij}\,(k) \leq 1$. These constraints are enforced by the BDI-C4 (single assignment) and BDI-C2 (completeness) contracts.

### D. Three-Level Parameter Space

Each architectural level operates with its own parameter space, control loop frequency, and learning mechanism.

Level 1 (individual, 10-50 ms cycle) is characterized by plastic synaptic weights $W_i(t)$ , Hebbian learning rate $\eta_H$ , neuromodulation signal $M_i(t)$ from Level 2, and frozen safety synapses $\mathcal{S}_{\text{frozen}}$.

Level 2 (tactical, 0.1-1 s cycle) involves GNN parameters $\theta_{\text{GNN}}$ with Graph Attention architecture, MARL policy parameters $\theta_{\text{MARL}}$ optimized by PPO, coordination embedding dimension $d_{GNN}$, and policy update bound $\Delta_{max}$.

Level 3 (strategic, 1-10 s cycle) manages meta-learning parameters $\theta_{\text{meta}}$ with MAML, a digital twin environment for safe strategy evaluation, a BDI reasoning engine with bounded deliberation $H_D$, and a library of meta-knowledge $M$ that grows with mission experience.

## III. THREE-LEVEL LEARNING ARCHITECTURE

The proposed architecture assigns qualitatively different learning mechanisms to each hierarchical level. This design is motivated by three observations. First, biological intelligence exhibits a clear hierarchy: spinal reflexes (fast, local, hardwired), motor skills (coordinated, learned, moderately adaptive), and reasoning (slow, strategic, highly flexible). Second, the requirements of SAR operations span this entire spectrum: collision avoidance must be reflexive, formation maintenance requires coordination, and mission replanning demands strategic thinking. Third, using different learning types at different levels provides natural graceful degradation: when higher levels fail, lower levels continue operating independently.

### A. Level 1: Hebbian Neuroplasticity

The lowest level provides individual agent adaptation through Hebbian neuroplasticity, a biologically inspired learning mechanism that modifies synaptic weights based on local activity correlations without requiring an external training signal. Each agent $\mathcal{A}_i$ possesses a plastic neural controller whose synaptic weights are updated online according to a modulated Hebbian rule:

$$\Delta w_{ij}(t) = \eta_{\mathrm{H}} \cdot M_i(t) \cdot \left(x_i(t) \cdot y_j(t) + \alpha \cdot y_j(t) \cdot [r(t) - \bar{r}]\right)$$

where $x_i$ and $y_j$ are pre- and post-synaptic activations, $M_i(t)$ is a neuromodulatory signal from Level 2 that gates plasticity based on the tactical context, $\alpha$ weights the reward-modulated component, and $\bar{r}$ is a running average of recent rewards. This formulation combines classical Hebbian learning (correlation-based) with reward-modulated plasticity (reinforcement-sensitive), following the differentiable plasticity framework of Miconi et al.

An important consequence of this mechanism is controller divergence: agents with identical initial weights $W_i(0) = W_0$ develop distinct behavioral profiles through exposure to different sensory experiences, creating heterogeneous swarm behavior from homogeneous initialization. Formally, for two agents i and j exposed to different sensory streams, $\| W_i(T) - W_j(T) \|$ grows with the divergence of their sensory histories. This behavioral diversity improves robustness of the swarm by producing natural specialists adapted to different terrain types, lighting conditions, and obstacle configurations.

The computational and communication profile of Hebbian plasticity matters for Level 5 autonomy, formalized by contract NP-C3: $\mathrm{Compute}(\Delta w) = O(|\mathrm{synapses}|)$, $\mathrm{Communication} = 0$. The controller adapts using only local sensory information and local computation, requiring zero communication with other agents or with higher architectural levels. This means that even under complete communication loss (electronic warfare conditions, GPS denial), each agent continues to adapt its behavior to local conditions. This is the foundation of graceful degradation in the architecture.

Safety under plasticity is guaranteed by contract NP-C2 (safety-compatible plasticity): a subset of synapses $\mathcal{S}_{\mathrm{frozen}}$ is excluded from Hebbian modification ($\Delta w_s(t) = 0$ for $s \in \mathcal{S}_{\mathrm{frozen}}$). These frozen synapses encode critical safety responses (collision avoidance, geofence compliance, emergency return-to-base) that remain invariant regardless of how other weights evolve through experience. This creates a protected behavioral core that survives any amount of adaptation.

### B. Level 2: MARL with GNN Coordination

The tactical level coordinates agent groups through multi-agent reinforcement learning integrated with graph neural networks and Behavior Trees. This level operates at a 0.1–1 second cycle and is responsible for formation maintenance, task allocation within groups, communication routing, and coordinated maneuvers.

#### 1) Graph Neural Network Architecture

The GNN aggregates information from neighboring agents using a Graph Attention Network (GAT) architecture. The communication graph $G(t) = (V, E(t))$ is dynamic: vertices $V$ correspond to agents, and edges $E(t)$ are defined by physical proximity and communication range $R_{comm}$. The GNN computes coordination embeddings through multi-head attention:

$$h_i^{(l+1)} = \sigma\left(\sum_{j \in \mathcal{N}(i)} \alpha_{ij}^{(l)} \, W^{(l)} \, h_j^{(l)}\right)$$

where $h_i^{(l)}$ is the embedding of agent $i$ at layer $l$, $\mathcal{N}(i)$ is its neighborhood in $\mathcal{G}(t)$, $\alpha_{ij}^{(l)}$ are attention coefficients computed via a learned compatibility function, $W^{(l)}$ is a shared weight matrix, and $\sigma$ is a nonlinear activation.

The output $h_i^{(L)}$ is the coordination embedding passed to the MARL policy of each agent about its neighborhood state.

A critical property of the GNN is permutation equivariance (contract GNN-C3):

$$\mathrm{GNN}(\Pi \mathbf{X}) = \Pi \cdot \mathrm{GNN}(\mathbf{X})$$

for any permutation matrix $\Pi$, where $\mathbf{X}$ is the matrix of all agent embeddings. This ensures that coordination results are independent of agent numbering, which has a practical consequence: adding or removing agents during a mission does not require retraining the GNN. The GNN architecture also provides bounded approximation error (contract GNN-C1): $\| \hat{F}_{gj} - F_{gj} \|_\infty \leq \varepsilon_{\mathrm{GNN}}$, bounding the gap between the true progress function and its neural approximation.

#### 2) MARL Policy Optimization

The MARL component employs Proximal Policy Optimization (PPO) with a credit assignment mechanism based on QMIX-style monotonic value factorization. The joint value function $Q_{\mathrm{tot}}$ is decomposed as a monotonic function of individual Q-values: $\partial Q_{\mathrm{tot}} / \partial Q_i \geq 0$, which guarantees that improving one agent's policy does not decrease the collective reward. PPO's clipping mechanism ensures bounded policy updates:

$$\left\|\pi_i^{(k+1)} - \pi_i^{(k)}\right\|_{\mathrm{TV}} \leq \Delta_{\max}$$

(contract MARL-C1), which is critical for stability in the multi-agent setting where each agent's environment includes the policies of all other agents.

The safety layer separation (contract MARL-C2) ensures that MARL policies operate exclusively on leaf nodes of the mission branch in the Behavior Tree, with no programmatic interface to safety actions. This is enforced architecturally: the policy network output is masked by a safety layer that sets

$$\pi_i(a \mid s) = 0 \quad \text{for all } a \in \mathcal{A}_{\text{unsafe}}(s)$$

Violations are detectable by static analysis of the network interface.

*3) Behavior Tree Integration*

The Behavior Tree (BT) provides the execution framework that integrates safety guarantees with learned policies. Contract BT-C1 mandates that the BT root has a selector structure with the safety branch as its first child: Safety Selector -> [SafetyBranch, MissionBranch]. Contract BT-C2 requires that safety interception occurs within one tick: any emergency signal triggers the safety response before the mission branch is evaluated. This structure ensures that no learned policy can override safety decisions, providing the formal safety guarantee.

### C. Level 1-2 Integration: Embedding Channel

The integration between Levels 1 and 2 is realized through the embedding channel: the output of each agent's neuroplastic controller becomes part of its node embedding in the GNN. Formally, $h_i^{(0)} = [\text{obs}_i;\ \text{emb}_i(W_i(t))]$, where $\text{obs}_i$ is the raw observation and $\text{emb}_i$ is a function of the plastic weights. This creates a direct influence pathway: individual adaptation modifies embeddings, which modify coordination through the GNN. The stability of this channel is bounded by the Lipschitz chain: $\| \pi_i(\text{emb}_{t+1}) - \pi_i(\text{emb}_t) \| \leq L_\pi \cdot L_e \cdot \Delta_{\text{NP}}$, where $L_\pi$ and $L_e$ are Lipschitz constants of the policy and embedding functions, and $\Delta_{\text{NP}}$ is the plasticity bound from NP-C1. Coordination stability requires $L_\pi \cdot L_e \cdot \Delta_{\text{NP}} \leq \Delta_{\max}$ (MARL-C1).

### D. Level 3: Meta-Learning with BDI Reasoning

The strategic level employs Model-Agnostic Meta-Learning (MAML) integrated with a Belief-Desire-Intention (BDI) architecture and a digital twin of the swarm. This level operates at a 1-10 second cycle and is responsible for mission-level strategy selection, task priority adjustment, and adaptation to novel operational scenarios.

*1) MAML-Based Strategy Adaptation*

MAML optimizes the initial parameter point $\theta$ such that a small number $K_{\text{inner}}$ of gradient steps on new task data yields an effective strategy:

$$\theta^* = \underset{\theta}{\text{argmin}} \sum_{\mathcal{T}_j \sim p(\mathcal{T})} \mathcal{L}_{\mathcal{T}_j} \left( \theta - \alpha \nabla_\theta \mathcal{L}_{\mathcal{T}_j}(\theta) \right)$$

where $\mathcal{T}_j$ are tasks sampled from the distribution $p(\mathcal{T})$, $\mathcal{L}_{\mathcal{T}_j}$ is the loss function for task $\mathcal{T}_j$, $\nabla_\theta$ is the gradient with respect to $\theta$, and $\alpha$ is the inner learning rate. The convergence of MAML has been established by Fallah et al. for strongly convex, Lipschitz-gradient loss functions at a rate of $O(1/\sqrt{T})$. In our architecture, the digital twin is the computational environment where MAML generates the task distribution $\mathcal{D}_{\text{task}}$ for the outer optimization loop, evaluating candidate strategies through accelerated simulation (up to 1000x real-time) without risk to the real swarm.

Contract ML-C1 (bounded adaptation) bounds the total adaptation time:

$$T_{\text{adapt}} = K_{\text{inner}} \cdot T_{\text{DT}} + T_{\text{deploy}} \leq T_{\text{critical}}$$

where $K_{\text{inner}}$ is the number of inner MAML gradient steps, $T_{\text{DT}}$ is the time per digital twin step, $T_{\text{deploy}}$ is the strategy propagation time, and $T_{\text{critical}}$ is the maximum acceptable reaction time. Typical values are $K_{\text{inner}} = 5$, $T_{\text{DT}} = 0.5$ s, $T_{\text{deploy}} = 2$ s, $T_{\text{critical}} = 10$ s. Contract ML-C2 (monotonic meta-improvement) ensures that $\mathbb{E}[K_{\text{inner}}^{(n+1)}] \leq \mathbb{E}[K_{\text{inner}}^{(n)}]$: with each successive novel situation, the swarm requires fewer adaptation steps, formalizing the notion that the swarm becomes more intelligent over time.

*2) BDI Reasoning Engine*

The BDI component provides explainable high-level reasoning through the classical Belief-Desire-Intention cycle. Beliefs are maintained about the environment state, victim locations, agent capabilities, and threat assessments, updated through sensor fusion and inter-agent communication. Desires encode mission objectives with priorities (coverage completion, victim rescue, asset preservation). Intentions are committed plans selected through a deliberation process that combines symbolic reasoning with neural evaluation.

The BDI engine governs the meta-learning process through five contracts. BDI-C1 (bounded desires) limits the desire buffer to $D_{\max}$ entries, preventing computational explosion. BDI-C2 (completeness) ensures that every ground task is considered during planning. BDI-C3 (belief consistency) bounds belief divergence from the true state. BDI-C4 (single assignment) enforces one-to-one agent-task mapping. BDI-C5 (replanning trigger) activates replanning when progress stalls or new tasks arrive. Together, these contracts ensure that the symbolic reasoning layer maintains consistency with the mathematical model variables.

*3) Digital Twin as Pre-Filter*

The digital twin acts as a safety pre-filter for strategy evaluation. Before any strategy selected by the meta-learning process is deployed to the real swarm, it is evaluated in the digital twin environment. The digital twin simulates the current mission state, agent configurations, and environmental conditions, allowing rapid assessment of strategy viability. This pre-filtering mechanism is not itself a formal safety guarantee (which is provided by BT-C1/BT-C2 and NP-C2), but it significantly reduces the probability of deploying suboptimal strategies to the operational swarm.

## IV. Mathematical Framework

### A. Progress Dynamics

The central equation of the mathematical model governs how tasks advance toward completion:

$$p_j(k+1) = \min\left(1,\ p_j(k) + \sum_g F_{gj}\left(y_{gj}(k)\right)\right)$$

where $p_j(k) \in [0,1]$ is the progress of task $j$ at step $k$, $F_{gj}$ is the progress function of group $g$ for task $j$, and $y_{gj}(k)$ is the allocated resource. The $\min(1, \ldots)$ operator ensures that progress remains in $[0,1]$. The progress is guaranteed to be monotonically non-decreasing (since $F_{gj} \geq 0$ by axiom F4) and bounded (by the min operator), providing the safety invariant for the progress domain.

### B. *Axioms of Progress Functions*

The abstract optimization theory requires that progress functions $F_{gj}: \mathbb{R}_{\geq 0} \rightarrow [0,1]$ satisfy four axioms:

(F1) $F_{gj}(0) = 0$, meaning zero resource yields zero progress;

(F2) $F_{gj}$ is monotonically non-decreasing;

(F3) $F_{gj}$ is concave;

(F4) $F_{gj}$ is continuous and non-negative.

The concavity axiom (F3) is the most consequential: it models diminishing returns, where each additional unit of resource produces less additional progress than the previous one. This property is physically natural for all five SAR task types and is the mathematical foundation for the optimality of greedy resource allocation.

- Constructive Progress Functions for SAR

Unlike prior work that assumes progress functions exist, we derive explicit parametric forms for all five SAR task types. Unlike existential assumptions common in optimization literature, each function is derived from a physical model of the corresponding process and formally verified to satisfy axioms (F1)-(F4).

Coverage: $F_{gj}^{\text{cov}}(y) = 1 - \exp\left(-\frac{s_g \cdot y}{A_j}\right)$, where $A_j$ is the area to be covered and $s_g$ is the effective scanning area per resource unit. Derived from a stochastic covering model: each agent-step independently scans a fraction $s_g/A_j$ of the zone, and overlaps follow a random covering process. The function is strictly concave ($F'' < 0$) and infinitely differentiable.

Verification: $F_{gj}^{\text{ver}}(y) = \frac{\min(K_j,\, y/\gamma_{gj})}{K_j}$, where $K_j$ is the number of independent observations required for confirmation and $\gamma_{gj}$ is the resource cost per observation. Derived from sequential hypothesis testing: each observation reduces uncertainty by a fixed amount until $K_j$ confirmations are achieved. The function is concave (piecewise linear with non-increasing slopes) but not strictly concave.

Tracking: $F_{gj}^{\text{trk}}(y) = \frac{\alpha_{gj} \cdot y}{\sigma_j^2 + \alpha_{gj} \cdot y}$, where $\sigma_j^2$ is the initial position uncertainty and $\alpha_{gj}$ is the sensor information rate. Derived from Kalman filtering: each measurement reduces the position variance, but with diminishing returns as the estimate improves. Strictly concave.

Delivery: $F_{gj}^{\text{del}}(y) = \min\left(1,\, \frac{v_g \cdot y}{d_j}\right)$, where $v_g$ is the group speed and $d_j$ is the delivery distance. A linear progress model: each resource unit advances the delivery by a fixed distance. Concave (piecewise linear) but not strictly concave.

Relay**:** $F_{gj}^{\text{rel}}(y) = \left(\frac{\min(L_j,\, y/\rho_{gj})}{L_j}\right)^{\theta_j}$, where $L_j$ is the number of relay links required, $\rho_{gj}$ is the resource cost per link, and $\theta_j \in (0,1]$ is a concavity parameter. Models the construction of a multi-hop relay chain with complementarity effects. Strictly concave when $\theta_j < 1$.

### C. *Optimization Objective*

The single-step optimization problem maximizes the weighted sum of progress increments: $\max \sum_j w_j \cdot F_j(y_j)$ s.t. $\sum_j y_j \leq B,\ y_j \geq 0$. The existence of an optimum follows from the Weierstrass theorem (continuous function on a compact set). The concavity of $F_j$ ensures that the greedy algorithm (allocating discrete resource units to the task with the highest marginal bid) achieves a $(1 - 1/e)$ approximation ratio for the submodular maximization problem.

### D. *Decentralized Auction Mechanism*

The centralized optimization is realized through a decentralized auction. The budget B is discretized into $M = \lfloor B/\delta \rfloor$ atomic units, where $\delta$ is the discretization step. Each group computes marginal bids: $b_j(m) = w_j \cdot (F_j(m\delta) - F_j((m-1)\delta))$. By the concavity of $F_j$, the bid sequence is non-increasing (diminishing returns), which is what makes greedy allocation near-optimal. The Tree-Max algorithm executes this auction in a decentralized manner over the communication graph, achieving the same allocation as the centralized greedy with a delay bounded by the graph diameter.

## V. CONTRACTUAL FRAMEWORK AND FORMAL GUARANTEES

### A. *Architecture of 22 Contracts*

The architecture rests on 22 architectural contracts organized across six components. Each contract is a verifiable specification consisting of a precondition (Pre), an invariant (Inv), and a postcondition (Post), following the design-by-contract methodology. Unlike abstract mathematical assumptions, each contract has a constructive realization through an architectural mechanism that can be tested.

BDI Contracts (BDI-C1 through BDI-C5):BDI-C1 bounds the desire buffer to $D_{\max}$ entries. BDI-C2 ensures planning completeness: every ground task is assigned when capable agents exist. BDI-C3 bounds belief divergence from the true state. BDI-C4 enforces single agent-task assignment through the commitment strategy. BDI-C5 triggers replanning when progress stalls beyond a threshold.

Behavior Tree Contracts (BT-C1 through BT-C3):BT-C1 mandates safety-first tree structure: the root selector evaluates the safety branch before the mission branch. BT-C2 requires one-tick safety interception: emergency signals trigger immediate response. BT-C3 ensures mission-safety separation: learned policies operate only on mission branch leaf nodes, with no interface to safety actions.

GNN Contracts (GNN-C1 through GNN-C3): GNN-C1 bounds the approximation error: $\| \hat{F}_{gj} - F_{gj} \|_\infty \leq \varepsilon_{\text{GNN}}$. GNN-C2 preserves concavity: the approximator is implemented as an Input Convex Neural Network (ICNN) with architectural constraints on weights that guarantee concavity by construction,

not by verification on each input. GNN-C3 ensures permutation equivariance.

MARL Contracts (MARL-C1 through MARL-C6): MARL-C1 bounds policy updates to $\Delta_{\max}$ in total variation distance. MARL-C2 separates safety from learning through the policy mask. MARL-C3 ensures approximately monotonic improvement: $J(\pi^{(k+1)}) \geq J(\pi^{(k)}) - \varepsilon_{\mathrm{MARL}}$ . MARL-C4 bounds credit assignment error to $\varepsilon_{\mathrm{CA}}$. MARL-C5 ensures coordination efficiency: joint reward exceeds the sum of individual rewards minus coordination overhead. MARL-C6 bounds exploration rate $\varepsilon_{\mathrm{expl}}(k) \to 0$ as $k \to \infty$.

Neuroplasticity Contracts (NP-C1 through NP-C3): NP-C1 (bounded plasticity) limits weight change per step: $\| W_i(t+1) - W_i(t) \| \leq \Delta_{\mathrm{NP}}$ . This contract links Levels 1 and 2: bounded plasticity ensures that the MARL policy does not break due to rapid embedding changes. NP-C2 (safety-compatible plasticity) freezes safety synapses. NP-C3 (energy efficiency) guarantees $O(|\mathrm{synapses}|)$ computation and zero communication.

Meta-Learning Contracts (ML-C1 through ML-C2):ML-C1 (bounded adaptation) ensures $T_{\mathrm{adapt}} \leq T_{\mathrm{critical}}$ . ML-C2 (monotonic meta-improvement) ensures that the expected number of adaptation steps decreases with each new experience: $\mathbb{E}[K_{\mathrm{inner}}^{(n+1)}] \leq \mathbb{E}[K_{\mathrm{inner}}^{(n)}]$ . The proof relies on MAML convergence theory under $\mu$-strongly convex, $L$-Lipschitz gradient loss functions.

## B. *Six Classes of Formal Guarantees*

The main integration theorem (Theorem HAF) establishes that when all 22 contracts are simultaneously satisfied, the hybrid neuro-symbolic system preserves six classes of formal guarantees.

(G1) Safety:The system never enters a dangerous state. Guaranteed by BT-C1 (safety-first structure), BT-C2 (one-tick interception),MARL-C2 (safety layer separation), and NP-C2 (frozen safety synapses). This guarantee holds independently of the states of other contracts: safety is unconditional.

(G2) Budget Correctness:Resource allocation satisfies capacity constraints at all times. Via BDI-C4 (single assignment) and MARL-C5 ( coordination efficiency).

(G3) Optimality: The greedy allocation achieves a $(1 - 1/e)$ approximation ratio for submodular maximization. Via GNN-C1, GNN-C2 (concavity preservation), and BDI-C2 (completeness). The approximation gap with neural approximation is bounded by $\frac{2M \cdot w_{\max} \cdot \varepsilon_{\mathrm{GNN}}}{\min_j w_j}$.

(G4) Liveness:Every task with available capable agents eventually receives allocation. Via BDI-C5 (replanning trigger) and BDI-C3 (belief consistency).

(G5) Starvation-Freedom:No agent is indefinitely denied task participation. Ground tasks receive priority through the separation constant $\lambda$.

(G6) Inter-Level Consistency:Architectural abstractions (beliefs, intentions, action nodes, learned policies) correctly map to mathematical model variables. This guarantee is novel: it has no analog in the abstract optimization theory and arises only when binding the abstract model to a concrete architecture.

## C. *Constructive Verifiability*

The contractual approach differs from prior work in its constructive verifiability. Each contract maps to a specific test: BT-C1 is verified by inspecting the behavior tree configuration file (static, one-time verification). BT-C2 is verified by injecting an emergency signal and measuring response latency. MARL-C2 is verified by static analysis: the policy network has no programmatic interface to safety actions. GNN-C2 is verified by construction: the ICNN architecture with non-negative weight constraints guarantees concavity structurally. MARL-C1 is verified by inspecting the PPO clipping parameter. NP-C1 is verified by the weight update norm check. This replaces the traditional approach of assuming that $F_j$ is concave with the constructive statement that the ICNN architecture with weight constraints guarantees $\hat{F}_j$ is concave by construction.

# VI. Inter-Level Integration and Graceful Degradation

## A. *Integration Theorem*

Theorem HAF holds for a static architecture with fixed post-training parameters. In a real system, parameters change during mission execution through three learning mechanisms. At Level 1, Hebbian modification of synaptic weights violates progress function stationarity. At Level 2, MARL policy updates with $\Delta_{\max} > 0$ and $\varepsilon_{\mathrm{expl}} > 0$ introduce non-stationarity. At Level 3, MAML updates modify the meta-policy affecting all lower levels simultaneously.

The five new contracts (NP-C1-C3, ML-C1-C2) extend the static Theorem HAF to the dynamic setting. The full integration theorem for the dynamic architecture establishes that when all 22 contracts are satisfied, the system preserves the original six guarantees plus three additional ones: (G7) cognitive resilience (the swarm recovers coordination after transient disruptions within bounded time), (G8) graceful degradation (performance degrades proportionally to the number of failed levels, not catastrophically), and (G9) monotonic meta-improvement (adaptation quality does not deteriorate with experience). The quantitative gap between the static and dynamic cases is bounded by the plasticity rate and exploration parameters: at typical values, the approximation gap is approximately 0.5% for SAR missions.

## B. *Graceful Degradation Hierarchy*

The hierarchical design provides a natural degradation path under progressive failure. Under normal operations, all three levels are active: Level 3 performs meta-learning and strategic adaptation, Level 2 coordinates groups via MARL and GNN, and Level 1 provides continuous individual adaptation. When the swarm loses inter-group communication (e.g., under electronic warfare), Level 3 deactivates: the swarm falls back to the last known strategy, continuing with Level 2 coordination within connected subgroups and Level 1 individual adaptation.

When intra-group coordination fails, Level 2 degrades to individual policies learned during training. Each agent executes its pre-trained MARL policy with frozen parameters, operating independently but following the behavior patterns acquired during the coordination phase. Finally, under complete isolation (no communication, GPS denial, sensor degradation), Level 1 continues operating with zero communication, adapting through Hebbian plasticity alone. The agent falls back to $\sigma_{\text{fallback}}$, a basic Behavior Tree with the plastic controller, providing Level 1 autonomy. This is the architectural realization of NP-C3: zero-communication adaptation.

### C. *Swarm Meta-Cognition*

Swarm Meta-Cognition (SMC) is defined as a compositional property (not emergent) arising from the structured interaction of all three architectural levels. The distinction is architecturally significant: emergent properties arise unpredictably from component interactions and cannot be derived from specifications, while compositional properties can be formally traced to the interfaces between components.

### D. *SMC Architecture*

The SMC module operates by monitoring collective cognitive metrics: coverage rate, detection confidence, coordination efficiency, agent health distribution, and communication graph connectivity. These metrics are compared against expected performance baselines generated by the digital twin for the current operational context. When the performance gap exceeds a configurable threshold, SMC triggers a cognitive strategy switch through Level 3.

The available cognitive strategies form a discrete set $\Sigma = \{\sigma_1, \dots, \sigma_K\}$, where each strategy $\sigma_k$ specifies:

(a) the weight vector for task priorities,

(b) the GNN aggregation parameters,

(c) the Hebbian modulation gain, and

(d) the exploration rate for MARL. Strategy selection is formalized as a meta-optimization: $\sigma^* = \text{argmax}_{\sigma \in \Sigma} \mathbb{E}[R(\sigma) \mid s_{\text{current}}]$, where the expectation is evaluated through the digital twin.

### E. *Compositional Derivation*

The compositional nature of SMC can be formally traced.

Level 1 provides the raw behavioral diversity through controller divergence (the observation channel).

Level 2 provides the coordination metrics through GNN embeddings (the coordination channel).

Level 3 provides the strategy evaluation through the digital twin (the evaluation channel).

SMC combines these channels: it uses Level 2 metrics to detect performance degradation, Level 3 to evaluate alternative strategies, and Level 1 diversity to ensure that the new strategy has behavioral support. The correctness of each channel is guaranteed by the respective contracts (NP-C1-C3, MARL-C1-C6, ML-C1-C2), and the composition of correct channels yields a correct SMC module.

## VII. COMPARISON WITH EXISTING APPROACHES

### A. *Addressing Five Limitations of Existing HRL*

Limitation 1: Single-type learning.Existing approaches (FuN, h-QMIX, HSD, HTN-Enhanced MARL) use the same gradient-based mechanism at all hierarchy levels. Our architecture uses three qualitatively different learning types: Hebbian (Level 1, zero-gradient, local), MARL with PPO (Level 2, policy gradient, coordinated), and MAML (Level 3, meta-gradient, strategic). This matches the biological hierarchy where spinal reflexes, motor cortex, and prefrontal cortex employ fundamentally different computational mechanisms.

Limitation 2: Non-stationarity. In standard MARL, all agents learn simultaneously, creating a non-stationary environment that violates convergence assumptions. Our NP-C1 contract bounds the rate of individual change ($\| \Delta W \| \leq \Delta_{\text{NP}}$), and the Lipschitz chain NP-C1 -> GNN-C1 -> MARL-C1 formally bounds how individual adaptation propagates to the multi-agent environment, providing a constructive stability guarantee absent from prior work.

Limitation 3: Neuro-symbolic integration.Existing cognitive architectures for swarms (ARCog-NET, NEUSIS, Agentic UAV) integrate symbolic and neural components without formal guarantees. Our 22-contract framework provides verifiable conditions under which the BDI symbolic layer and neural components maintain consistency (guarantee G6). Each symbolic construct (belief, desire, intention) maps to a specific mathematical variable with bounded approximation error.

Limitation 4: Safety in hybrid systems.No existing MARL architecture provides constructive safety guarantees for hybrid neuro-symbolic systems. Our safety guarantee (G1) is unconditional: it holds regardless of the states of other guarantees, through the BT-C1/BT-C2/NP-C2 contract chain that ensures safety even if learning diverges or communication fails.

Limitation 5: Individual-collective coupling. The relationship between individual agent adaptation and collective performance is not formalized in existing work. Our NP-C1/MARL-C1 Lipschitz chain provides the first formal bound on propagation: individual plasticity with bound $\Delta_{\text{NP}}$ results in collective policy change bounded by $L_\pi \cdot L_e \cdot \Delta_{\text{NP}}$, enabling quantitative analysis of the individual-to-swarm influence pathway.

### B. *Comparison with Specific Architectures*

Compared to graph-based MARL approaches (MAAC, HGAT, RODE), our architecture adds two critical dimensions: the Hebbian layer for zero-communication adaptation and the meta-learning layer for strategic reasoning. While RODE introduces role-based decomposition, it lacks formal safety guarantees and the ability to adapt to unseen situations without retraining. Compared to VLA models for UAVs (UAV-VLA, AutoFly, VLA-AN), which operate at the single-agent level, our architecture addresses multi-agent coordination, which remains an open problem for VLA integration.

## VIII. DISCUSSION AND FUTURE DIRECTIONS

### A. *Theoretical Limitations*

The architecture has well-defined theoretical boundaries. The MAML convergence guarantees assume strongly convex loss functions, which may not hold for all SAR scenarios involving highly non-convex reward surfaces. The six guarantees of Theorem HAF hold for the static architecture;

extending them to the fully dynamic case requires the five new contracts, and the resulting bounds depend on the plasticity and exploration parameters being sufficiently small. The constructive verifiability of contracts relies on architectural invariants that must be maintained during system updates and reconfiguration.

The model assumes that progress functions can be accurately estimated from physical parameters. In practice, environmental uncertainty (weather, terrain, dynamic obstacles) introduces noise in progress estimation. The bid noise bound $\eta$ formalizes this uncertainty and its impact on optimality (the approximation ratio degrades gracefully with $\eta$), but the calibration of progress function parameters remains an empirical challenge that requires mission-specific validation.

### B. Open Problems

Several research directions emerge from this work. First, PAC-Bayes bounds for online meta-learning would provide probabilistic guarantees for the adaptation quality in non-stationary environments. Second, scalable GNN architectures for swarms exceeding 100 agents require hierarchical graph attention with sparse communication to maintain real-time performance. Third, integration of Vision-Language-Action (VLA) models would enable natural-language interaction with the swarm, creating a human-swarm interface at Level 3. Fourth, formal verification of Rules of Engagement (ROE) compliance through Petri net models would address the ethical and legal constraints of autonomous systems.

The possibilistic extension of BDI reasoning presents a particularly promising direction. Current belief maintenance in the BDI engine uses probability distributions, but SAR operations often involve deep uncertainty (unknown unknowns) better modeled by possibility theory. Integrating possibilistic reasoning with meta-learning would enable the swarm to distinguish between situations that are rare but understood (probability) and situations that are genuinely novel (possibility). The extension to swarm-of-swarms coordination (multiple autonomous swarms cooperating across domains) requires hierarchical meta-cognition beyond the single-swarm SMC presented here.

### C. Roadmap

The near-term priorities (0-2 years) focus on simulation validation with hardware-in-the-loop testing, scalable GNN implementation,LLM-VLA interface prototyping, and edge inference optimization. The medium-term priorities (2-5 years) include flight testing with 10-30 UAVs, Petri net verification of ROE compliance, non-stationary meta-learning extensions, adaptive autonomy levels for human-swarm teaming, and neuromorphic computing integration for energy-efficient inference. The long-term vision (5-15 years) encompasses possibilistic BDI, multi-domain swarm-of-swarms, quantum-accelerated optimization (QAOA for auction mechanisms), and continual learning with formal guarantee preservation across missions.

## IX. Conclusion

This paper presented a three-level hierarchical learning architecture for autonomous UAV swarms in search-and-rescue operations. The architecture addresses five fundamental limitations of existing hierarchical RL approaches by assigning qualitatively different learning mechanisms to each level: Hebbian neuroplasticity for individual adaptation (Level 1), MARL with GNN coordination for tactical operations (Level 2), and MAML-based meta-learning with BDI reasoning for strategic decision-making (Level 3). This design mirrors the biological progression from reflexes through learned skills to deliberative reasoning.

The formal contractual framework of 22 architectural contracts (17 for the static architecture plus 5 for the dynamic learning setting) provides six classes of guarantees (safety, budget correctness, optimality, liveness, starvation-freedom, inter-level consistency) with three additional guarantees for the dynamic case (cognitive resilience, graceful degradation, monotonic meta-improvement). Each contract has a constructive realization through a testable architectural mechanism, distinguishing this work from approaches based on abstract mathematical assumptions.

Five constructive progress functions for SAR task types (coverage, verification, tracking, delivery, relay) bridge the gap between abstract optimization theory and concrete operational scenarios. The concept of Swarm Meta-Cognition (SMC) as a compositional property enables the swarm to monitor its own cognitive state and switch strategies without emergent unpredictability.

The architecture provides natural graceful degradation under progressive communication loss: from full strategic reasoning through coordinated execution to autonomous individual adaptation, which allows continued operation in contested electromagnetic environments. Future work focuses on experimental validation through simulation with realistic SAR scenarios, progressive deployment to hardware-in-the-loop testing, and extension of the contractual framework to non-stationary environments and multi-domain operations.

## X. Acknowledgment

This work is part of the research program on mathematical foundations of collective intelligence in hybrid multi-agent systems for search-and-rescue operations (Erasmus+ project No. 2023-2-PL01-KA220-HED-0001179445 “The Transferable Training Model – The Best Choice for Training IT Business Leaders”, TransLeader).

## AUTHORS' BACKGROUND

| Your Name | Title* | Research Field | Personal website |
| --- | --- | --- | --- |
| Oleksii Bychkov | Dr.Science, full professor | Computer Science | Pst.knu.ua |